\relax
\documentclass[letterpaper]{article} 
\usepackage{aaai22}  
\usepackage{times}  
\usepackage{helvet}  
\usepackage{courier}  
\usepackage[hyphens]{url}  
\usepackage{graphicx} 
\usepackage{natbib}  
\usepackage{caption} 
\DeclareCaptionStyle{ruled}{labelfont=normalfont,labelsep=colon,strut=off} 
\usepackage{algorithm}
\usepackage{algorithmic}
\usepackage{amsfonts}
\usepackage{amsmath}
\usepackage{booktabs}
\usepackage{newfloat}
\usepackage{listings}
\floatstyle{ruled}
\newfloat{listing}{tb}{lst}{}
\floatname{listing}{Listing}
\title{Deploying clinical machine learning? Consider the following...\thanks{The authors thank John Chen, James Hillis, and Bernardo Bizzo for their support and helpful discussions.}}

\author {
    Charles Lu,\textsuperscript{\rm 1, 3} 
    Ken Chang,\textsuperscript{\rm 2, 3} 
    Praveer Singh,\textsuperscript{\rm 2} 
    Stuart Pomerantz,\textsuperscript{\rm 1} 
    Sean Doyle,\textsuperscript{\rm 1} 
    Sujay Kakarmath,\textsuperscript{\rm 1} 
    Christopher Bridge,\textsuperscript{\rm 1, 2} 
    Jayashree Kalpathy-Cramer\textsuperscript{\rm 1, 2}
}
\affiliations {
    \textsuperscript{\rm 1} Massachusetts General Hospital, Boston, MA\\
    \textsuperscript{\rm 2} Harvard Medical School, Boston, MA\\
    \textsuperscript{\rm 3} Massachusetts Institute of Technology, Cambridge, MA\\
}

\usepackage{bibentry}

\begin{document}

\maketitle

\begin{abstract}
     Despite the intense attention and considerable investment into clinical machine learning research, relatively few applications have been deployed at a large scale in a real-world clinical environment. 
     While research is essential in advancing the state-of-the-art, translation is equally important in bringing these techniques and technologies into a position to impact healthcare ultimately. 
     We believe a lack of appreciation for several considerations is a significant cause for this discrepancy between expectation and reality.
     To better characterize a holistic perspective among researchers and practitioners, we survey several practitioners with commercial experience developing machine learning tools for clinical deployment.
     Using these insights, we identify several main categories of challenges to design better and develop clinical machine learning applications. 
\end{abstract}


\section{Introduction}
\label{sec:intro}
    After the COVID-19 pandemic, hundreds of AI papers were published, but few were clinically practical or useful~\cite{born2021role,roberts2021common}.
    Even large technology companies have recently encountered unexpected difficulties in deploying state-of-the-art research into actual clinical products for healthcare applications~\cite{vincent2021google,ibmfail}.
    We believe there is a disconnect between the clinical machine learning (CML) research community and delivered clinical impact from the translation of useful CML tools~\cite{10.3389/fdgth.2021.594971}.
    While there have been extensive claims of AI performance being equivalent or superior to human doctors within the research literature~\cite{liu2019comparison}, emerging clinical studies reveal contradictions in those claims of purported super-human performances~\cite{Freemann1872,zech2018variable, Voter}.
    
    Machine learning (ML) in the healthcare domain faces numerous challenges, such as immense difficulty in acquiring and annotating large amounts of medical data, recruiting the necessary clinical expertise in validating models, and integration into existing clinical workflows and infrastructure~\cite{challenges, 10.1001/jama.2018.11029}. 
    Developing CML software is considered incredibly challenging compared to traditional software systems~\cite{design, DBLP:journals/corr/abs-2003-07678, practical}.
    Furthermore, sub-fields of healthcare such as oncology and dentistry have even more nuanced challenges to adopting AI to their respective specialties~\cite{cancer,  dentistry}. 
    Others have attempted to operationalize these challenges into a formal specification; \citet{pmlr-v136-oala20a} apply the ITU/WHO FG-AI4H framework to audit several case studies of AI applications for diagnostic retinopathy, Alzheimer's diagnosis, and cytomorphologic classification for leukemia.
    
    To gain a more holistic understanding of the challenges in CML translation, we survey researchers and clinicians and distill their insights into better practices.
    The contributions of our survey complement other works, such as the perspective by~\cite{Wiens2019DoNH} by providing empirical insights grounded in actual interviews with working CML practitioners.
    By enumerating these considerations explicitly, we hope to promote and drive more targeted CML research that is better situated in the clinical context and facilitate the translation of research into real clinical benefit.
    
\begin{table}
    \scriptsize
    \centering
    \begin{tabular}{c|c|c}
        \textbf{Role} & \textbf{Years of experience} & \textbf{Considerations}\\
        \hline
        Machine learning scientist & 5 & 2.1, 3.2, 4.1, 4.2\\
        Machine learning scientist & 10 & 1.2, 2.2, 3.1, 3.3\\
        Software engineer & 30 & 2.4, 3.2, 3.3, 4.1 \\
        Clinical project manager & 5 & 1.1, 1.3, 2.1, 4.2\\
        Radiologist & 25 & 1.3, 2.3, 4.3\\
        Neurologist & 10 & 1.3, 2.3\\
        Clinical researcher & 5 & 2.1, 4.2\\
    \end{tabular}
    \caption{Role and (approximate) experience level of surveyed participants and their (general) contributions to specific considerations.}
    \label{tab:field}
\end{table}

\section{Considerations}
    First, we detail the survey methodology. 
    We conducted several semi-structured interviews with clinical and technical practitioners from the Center of Clinical Data Science at the Massachusetts General Hospital and Brigham and Women's Hospital.
    Participants were chosen based on their prior experience developing and deploying CML products in collaboration with industry partners for stroke, emergency medicine, and radiology applications.
    In each session, we asked several open-ended questions to identify reoccurring themes and asked probing follow-up questions tailored to the background and expertise of each participant. 
    Examples of questions included:
    \begin{quote}
        \small
        What are the major challenges you encountered when deploying clinical machine learning?
    \end{quote} 
    and
    \begin{quote}
        \small
        What would you have done differently if you had to start this project over?
    \end{quote}
    
    We analyzed the responses, made broad thematic categories, and synthesized specific considerations that would generally apply to most CML contexts.
    Table \ref{tab:field} summarizes the interviewed participants' roles, experiences, and contributions. While the considerations collected should not be taken as an exhaustive list, we believe they can still serve as a fundamental and approachable checklist of real-world considerations for most ML engineers and researchers working towards clinical application. We group these considerations into four broad areas: clinical context, clinical validation, deployment, and monitoring.
    
    \subsection{1) Clinical context}
        The clinical context is often overlooked in CML research in favor of demonstrating high performance or technical novelty on benchmark datasets ~\cite{yan2018deeplesion,codella2019skin, abramoff2016improved}.
        Commercial CML systems designed for clinical practice also meet unexpected challenges and friction in deployment and integration~\cite{ibmfail, DBLP:journals/corr/abs-2101-01524}.
        For example,~\citet{DBLP:journals/corr/abs-2101-01524} report the tensions between the AI clinical decision support system and the reality of clinical practice in rural China.
        \subsubsection{1.1 Identify pain points}
            CML research focusing on automated prediction for critical diseases may often miss lower-hanging fruit with a greater potential utility to the clinical workflow. 
            Specifically, tools that are most likely to be appreciated improve the clinician's quality of life by decreasing administrative burden in routine clinical practice(e.g., replacing manual volume contouring with fully automatic 3D segmentation~\cite{macruz2022quantification}).
            In radiology, auto-generated report templates and speech dictation are conveniences that radiologists have widely adopted because they decrease friction and interpretation time in the overall workflow. 
            
            Some of the best product ideas come from clinicians and vendors of clinical software. 
            Clinicians are well-positioned to estimate the potential value generated from CML tools that directly address experienced pain points. 
            Industry vendors will have already assessed the market value and feasibility of software integration costs into existing clinical infrastructure. 
            The focus should be on developing tools for a particular stage or aspect of the clinical workflow (such as triage, screening, or diagnosis), the associated downstream risks, and how they change or affect the existing care pathway.
            Close collaboration between researchers, clinicians, and industry vendors is often a wise route in CML research and development.
            
        \subsubsection{1.2 Assess feasibility}
            Many projects often underestimate the difficulty of realistic validation and deployment of CML models. 
            CML applications are uniquely challenging due to the need for clinical expertise to define cohort selection, annotate data, and evaluate model performance. 
            The lack of easily accessible and available data resources is a significant issue in CML. 
            In addition to maintaining patient privacy, annotating large amounts of sensitive medical data can be extremely difficult and more complicated than other kinds of data~\cite{gupta2023collaborative}.
            Inter-annotator and intra-annotator variability is a known issue; disagreements often arise due to differences in interpretation, training, and expertise of the annotator. 
            Consensus labeling requires multiple annotators, which places additional overhead and cost constraints.
            
            Clinical evaluation can also be subjective due to a lack of established evaluation standards. In such scenarios, performance metrics must be explicitly defined for the clinical application. 
            Furthermore, these assessment criteria may evolve under changing clinical guidelines or protocols.
            The exclusion criteria may need to be continuously refined as outliers are uncovered. 
            CML projects are an iterative development process requiring constant feedback between clinical and technical teams.
            
        \subsubsection{1.3 Promote clinical champions}
            All CML projects should have a dedicated clinical champion, preferably a practicing specialist in the area of the specific application. 
            The success of a project (even progressing past the initial development to the deployment testing) often depends on finding a ``clinical use-case fit''. 
            A clinical champion can continuously reaffirm the clinical utility, guide early development, and provide a path for later deployment.
            Ideally, a clinical champion will be an insider that can promote the project within complex social hierarchies within a hospital that would be impenetrable to outsiders. 
            Their dedicated buy-in often allows for prioritizing the project within a myriad of competing interests and helps in sourcing users to test CML prototypes.
            Especially for critical decision-making, trust is a major barrier to adoption ~\cite{DBLP:journals/corr/abs-2102-00593,lu2022improving}, and a clinical champion can help in defining what trust means to their colleagues, as well as assist in educating and training users of CML tools.
    
        \subsection{2) Clinical validation}
            Rigorous evaluation of many CML papers according to clinical standards is often lacking~\cite{ nagendran2020artificial}. 
            Historically, the ML research community has often focused on achieving state-of-the-art performance, while the medical community focused on demonstrating utility in broad clinical contexts and trials. 
            Some of these differences can be resolved using multi-site evaluation, choosing more clinically meaningful performance metrics, and performing user studies on end-to-end systems that simulate real-world clinical environments.
            
            \subsubsection{2.1 Multi-site evaluation}
                Generalizability across institutions is one of the biggest challenges to developing robust CML. A model might go from having an excellent performance at one site to having a considerably worse performance at another seemingly similar institution~\cite{chen2021deep, chang2020multi}. 
                This generalization gap can arise from factors such as differences in patient population, disease characteristics, or data acquisition settings~\cite{gupta2021addressing}.  
                This is why multi-site testing is crucial to assess and validate CML models honestly. 
                However, a study by~\cite {fdaeval} analyzed 130 FDA approvals of AI medical devices and found that 93 (72\%) did not publicly report evaluation on a multi-site evaluation set.
                
                A metastudy by~\citet{Freemann1872} finds that the evaluation procedures in many papers on CML for breast cancer screening are vulnerable to numerous biases in patient selection, ground truth reference standard, and human comparator evaluation -- ultimately, that inflating claims about CML being ``better than radiologists'' in detecting breast cancer. 
                34 out of 36 (94\%) CML systems were considered less accurate than a single radiologist, while all 36 were less accurate than the consensus of two or more radiologists. 
                We advocate testing CML on heterogeneous, multi-site datasets.
                Ideally, the gold standard evaluation should be a prospective study in clinical trials such as a multi-case multi-reader design with statistically meaningful endpoints~\cite{mcmr}.
                
            \subsubsection{2.2 Choose meaningful metrics}
                While many performance metrics overlap between ML and CML, traditional ML does not fully consider potential clinical benefits or demonstrated impact when deployed into real-world clinical workflows.
                A CML system to triage stroke patients might be evaluated for high sensitivity (true positive rate) as well as time advantage over standard-of-care, while a CML application to screen breast cancer risk might consider specificity (true negative rate) as well as callback rate~\cite{lu2022three,lu2021evaluating}.
                
                Meaningful evaluation metrics should align with the intended clinical application use case.
                For example, the goal of a CML application to detect sepsis is not merely the absolute rate of sepsis detection but to identify those cases of sepsis in which the clinician missed it; additionally, the CML must be able to detect sepsis early enough for there to be time for some clinically actionable intervention to impact care~\cite{goh2021artificial}.
                
                Also, despite the theoretical and pedagogical emphasis on binary classification, most tasks in medicine are not simple binary decisions but nuanced gradations that consider the clinical context and patient history. These grey zones of ambiguity and ambivalence should be recognized and deliberately acknowledged when operationalizing the problem into a more amendable form for CML.
                
            \subsubsection{2.3 Perform user studies}
                CML papers often ignore human factors and other social components in their evaluation. If the goal is to develop more useful CML systems for clinicians, more attention should be placed on the interaction between humans and CML applications~\cite{humanai}. 
                User studies with appropriate domain experts should participate in evaluating human-centered tools that empower clinicians~\cite{DBLP:journals/corr/abs-1902-02960,kiani2020impact,gaube2021ai}. 
                
                User studies can also identify and alleviate points of friction that hamper the successful adoption of CML applications.
                The more rigorous the user study, the more assurance can be had that the CML system will be used as intended (or at least that off-label uses can be discovered early). 
                Performing realistic user studies can also uncover small changes that substantially improve the user experience.
                For example, instead of outputting out uncalibrated probabilities from a black box model, semantically meaningful outputs can often be designed that better map onto user intuition (e.g., displaying ``mild'', ``moderate'', or ``severe'' instead of 73.2\% to reduce number fatigue and cognitive overhead). 
                Clear visualizations and thoughtful interface design can bring greater transparency that assists medical decision-making and facilitate trust in an algorithmic system~\cite{DBLP:journals/corr/abs-2001-05149}. 
        
            \subsubsection{2.4 Embed situated workflows}
                Integration of CML can also affect the flow of information. 
                For example, should the results of radiology CML be made available to the referring clinician? 
                What if there are multiple algorithm outputs, each of which may not be consistent with the report? 
                Which parts of the CML output belong in the patient's medical record? 
                There should be a way for the clinician to curate algorithmic results into clinically useful outputs. 
                
        \subsection{3) Deployment}
        \label{sec:deploy}
            Deployment in CML often has to follow stringent integration requirements with existing clinical infrastructure, which often supports other functions beyond that specific AI tool. 
            Practical implementation challenges of deploying into hospitals often necessitate considerable coordination between internal and external teams to set up orchestration and inference platforms (usually with interaction between third-party clinical software vendors). 
            
            \subsubsection{3.1 Development To Deploy}
                Decisions made during model development can have significant knock-on effects that ultimately affect the ease with which the model can be deployed~\cite{lu2020overview}. 
                Thus, many pain points in the model deployment process can be mitigated by developing the model with deployment in mind from the outset. 
                Relying on complex or slow data processing routines can significantly complicate deployment efforts. 
                Furthermore, choosing a modular codebase structure more easily allows for the re-use of code between the training and deployment stages.
        
            \subsubsection{3.2 Choose an integration layer}
                The difficulty of integration dramatically depends on where the CML application will be injected into the workflow. 
                For example, an application to detect motion artifacts might be deployed into the MRI scanner. In contrast, another application to triage stroke patients may be added directly to the clinical work-list software. Yet another critical finding is that notification applications for stroke may be integrated into a mobile device environment.
                Additionally, different inference contexts will have different requirements for the application. 
                An application deployed in the emergency room may require a faster inference time than an application running in the background on routine cases.
                
            \subsubsection{3.3 Embrace healthcare standards}
                The lack of support for standards often increases integration costs.
                Hospitals are often unaware of the importance of standards and fail to specify newer standards in their Request for Proposals (RFPs) when purchasing products.
                Embracing existing healthcare standards can help avoid proprietary APIs, which can lock users and hospitals into workflows that may be sub-optimal. 
                The Integrating the Healthcare Enterprise Initiative (IHE) has guidelines for integrating imaging models into clinical PACS~\cite{ihe-rad-air} through the use of existing standards such as DICOM Structured Reporting~\cite{dicom-sr}, DICOM Segmentation, and HL7. 
                Unfortunately, support for these standards is not yet widespread in enterprise software. 
                We contend that support for standards should be a key criterion hospitals use in their RFP process when purchasing systems to be as prepared as possible for future CML deployments.
                
            
        \subsection{4) Monitoring}
        \label{sec:monitor}
            Even after being deployed into clinical practice, CML models must be continuously monitored to record and audit predictions for safety (especially the side-effect of software changes and model updates).
            CML applications must consider traceability and consider how the system will be equipped to handle changing data streams while keeping track of predictions in a controlled and documented procedure.
            
            \subsubsection{4.1 Design for traceability}
                Who bears liability when CML (inevitably) makes an error and causes harm?
                Rather than conceal failures and shirk from reporting, we believe that the long-term interest of all parties lies in responsible disclosures (already widely practiced by computer security researchers) and reviews (similar to mortality conferences practiced by surgeons)~\cite{morbid}.
                The gradual accumulation of incident reports should alert manufacturers to failure modes, which might be more prone in specific risky subgroups or edge cases~\cite{evidencebased,biondetti2020name}. 
                Tracking CML predictions will also help to verify the intended use (as opposed to off-label use) and inform the development of proper mitigation strategies, such as user training or re-calibration of prediction thresholds. 
                
                Patterns of traceability can be very subtle. 
                If an algorithm has been part of the workflow for an extended period and has known failure modes -- when a new version of the algorithm is installed, it might be prudent to run both versions for a while and alert users if predictions on the new algorithm differ substantially from the old algorithm. 
                If the CML is from a different vendor, this task may be difficult if the CML output is not in a standardized format.
            
            \subsubsection{4.2 Beware of bias}
                Any ML model is susceptible to encoding and exacerbating existing disparities and biases from the data (during collection, selection, annotation, etc.) to algorithm development (design choices, training parameters, testing evaluation) to the framing of the application itself (purpose, reification, intended vs. actual usage)~\cite{barocas-hardt-narayanan, Green2018TheMI, DBLP:journals/corr/abs-1912-05511, DBLP:journals/corr/abs-2006-09663,biondetti2020name}. 
                As one recent study demonstrated on multiple medical datasets, CML models can discriminate patient race even in extremely down-sampled chest radiographs as small as 4x4 pixels, that would be unrecognizable to any human ~\cite{DBLP:journals/corr/abs-2107-10356}. 
                Other studies have shown that algorithms can have differential performance across subgroups, raising concerns surrounding equity and fairness~\cite{larrazabal2020gender,pierson2021algorithmic,lu2021evaluating,lu2022fair}.
                
                One clinician expressed the need for a platform with the ability to audit CML model for algorithmic bias across patient demographics, such as age and race, as well as acquisition metadata, such as time and version of model prediction~\cite{lu2021fair,lu2021evaluating}.
                Dashboards that interactively display different cohorts would allow clinicians more insight into how CML models perform in real-time.
                CML monitoring should flag any trends among sensitive subgroups, which, depending on the context, may be clinically warranted or not.
        
        
            \subsubsection{4.3 Expect distribution shift}
                We found consensus with other researchers for the need to consider non-stationarity in medical data~\cite{DBLP:journals/corr/abs-1806-00388}. 
                Populations, technology, and human behavior change over time. 
                Software will be modified, equipment will be replaced, and clinical guidance will be updated~\cite{ovadia2019trust,finlayson2020clinician}. 
                \citet{googleflu} studied concept drift in Google searches over a period that systematically overestimated flu incidence based on users' previous search behavior; predictive performance was observed to have deteriorated over time as their ML model became stale and outdated.
                
                Methods of detecting and characterizing distribution shift in clinical contexts should be a crucial area of continued research for CML deployment~\cite{rabanser2019failing, DBLP:journals/corr/abs-2103-11163,lu2022estimating}. 
                Interesting dynamics will emerge as CML systems begin to affect clinical decision-making, affecting patient outcomes, which in turn, feedback into population data that will be used in future CML systems~\cite{doi:10.1148/radiol.2020200038}.
                
            \subsubsection{4.4 Missing outputs}
                Once CML has become an established part of a clinical workflow, a clinician may begin to rely on its presence and expect default behaviors. 
                If the algorithm fails for whatever reason (e.g., network issue, data issue, algorithm issue), then a clinician may wait for the results believing that the CML output is in the queue. 
                Currently, there is no standard mechanism to report and handle the failure in which the prediction result will be unavailable, incomplete, or missing.
        
\section{Conclusion}
\label{sec:conclusion}
    We detail several essential considerations for the deployment of CML systems. 
    We hope these consideration promotes and drives more targeted research to address to situate CML into clinical application better and facilitates more translation of promising CML research into clinical practice to effect higher healthcare standards.

\bibliography{aaai22.bib}

\appendix


    

\end{document}